\documentclass[sn-mathphys]{sn-jnl}

\jyear{2021}%

\theoremstyle{thmstyleone}%
\theoremstyle{thmstyletwo}%

\theoremstyle{thmstylethree}%

\begin{document}

\title[Should Machine Learning Models Report to Us When They Are Clueless?]{Should Machine Learning Models Report to Us When They Are Clueless?}


 \author*[1]{\fnm{Roozbeh} \sur{Yousefzadeh}}\email{roozbeh.yousefzadeh@yale.edu}

 \author[2]{\fnm{Xuenan} \sur{Cao}}\email{xuenan.cao@yale.edu}


 \affil*[1]{\orgdiv{Yale Center for Medical Informatics and VA Connecticut Healthcare System}, \orgname{} \orgaddress{\street{} \city{New Haven}, \postcode{06510}, \state{CT}, \country{USA}}}

 \affil[2]{\orgdiv{Department}, \orgname{Organization}, \orgaddress{\street{Street}, \city{New Haven}, \postcode{06510}, \state{CT}, \country{USA}}}



\abstract{The right to AI explainability has consolidated as a consensus in the research community and policy-making. However, a key component of explainability has been missing: extrapolation, which describes the extent to which AI models can be clueless when they encounter unfamiliar samples (i.e., samples outside the convex hull of their training sets, as we will explain down below). We report that AI models extrapolate outside their range of familiar data, frequently and without notifying the users and stakeholders. Knowing whether a model has extrapolated or not is a fundamental insight that should be included in explaining AI models in favor of transparency and accountability. Instead of dwelling on the negatives, we offer ways to clear the roadblocks in promoting AI transparency. Our analysis commentary accompanying practical clauses useful to include in AI regulations such as the National AI Initiative Act in the US and the AI Act by the European Commission.
}




\maketitle

\section{Introduction}

A consensus has consolidated in the research community and policy-making about the right to reasonable explanations for people affected by decisions made by Machine Learning and Artificial Intelligence models \cite{wachter2018reasonable,coyle2020explaining}. In 2020, the National Artificial Intelligence Initiative Act in the United States recognized the need to improve the reliability of artificial intelligence methods. In 2021, the AI Act by the European Commission drafted a highly sophisticated product safety framework to rank and regulate the risks of AI-driven systems. Both acts hover above the key concern of the right to explanation. However, one fundamental element of the right to explanation has been neglected: extrapolation, which AI and ML models frequently perform. We propose that regulations incorporate articles requiring AI and ML models to report, for each decision or prediction that they make, whether they have extrapolated or not, and in which directions.

 
\section{Learning from data} 

AI and ML, broadly defined, is a set of mathematical methods automating the learning process. Using certain algorithms, a model learns from a training set (data on which the model is trained), then uses the learned phenomenon to make decisions and predictions in the world at large. In a medical setting, a model may learn from the clinical outcomes of a cohort of patients, and possibly predict with some accuracy for new patients that walk through the door of a hospital. It would be commonsensical for any health provider to inquire how a new patient compares with the cohorts of patients in the training set and whether the new patient's information falls within the range of information in the training set. Extrapolation is a mathematical concept describing just that. In an extreme case of extrapolation, a new patient could have some rare and complicated form of liver disease that the model has never seen before, and therefore, the model's output for this patient may not be reliable. If a nurse encounters a patient with features that he has never seen before, he may elevate the situation to an expert physician. A physician may also need to elevate certain cases to a committee of experts. One would expect, quite reasonably, a nurse or a physician to elevate such cases and seek further expertise. However, this procedure of escalating and reporting has so far eluded the attention of the research community and has been overlooked in regulations guiding the use of AI.

\section{Measuring extrapolation}

In math, there are well-defined algorithms for verifying whether a model is extrapolating, and if so, in which directions and dimensions. A training set, however small or large, forms a convex hull. Think of it as a dome. Any new sample, e.g., information about a new patient, will either fall within that convex hull or outside it. When a new data point is outside the convex hull of its corresponding training set, a model will need to extrapolate to process it. Conversely, the model would interpolate when a new data point is within the convex hull of its training set. The concept of convex hull dates back to at least Isaac Newton \cite{newton2008mathematical}. Extrapolation also has a rich literature in pure and applied mathematics \cite{brezinski2013extrapolation} and cognitive science and psychology \cite{yousefzadeh2021extrapolation}.



The concern of adding computational burden underscores the difficulty of determining whether a model has extrapolated. Our proposal shows that, instead of adding roadblocks, determining extrapolation adds a negligible computational burden while helping resolve issues of distrust. Transparency about extrapolation will increase the trust in using these automated systems. For example, physicians may not be willing to use automated systems in the medical setting unless the model provides adequate explanations for its recommendations and abstains from making decisions when it is clueless. Increased transparency can help gain the confidence of expert physicians to use these models in the first place.

\section{Right to explanation}

Whether a model has extrapolated is a piece of information lying at the heart of the right to explanation. In automated decision-making, if a model is making vital decisions or predictions about a patient with features not similar to any sample it has seen before, the model should be mandated to report and elevate the case to human experts.

Consider, in a case of loan applications being decided by an automated system, extrapolation might happen for an applicant because she is an immigrant, relatively young, and very well educated, and the model has not seen any profile of an immigrant as young and educated in the training set. In such a case, the model might not make a sound decision, and it would be reasonable to have a loan officer look over the model's decision. If an automated process rejects a loan application while extrapolating, the model should report the direction and extent of extrapolation.

\section{Promoting transparency or keeping the extrapolation information hidden?}



We propose to make it transparent whether models extrapolate or not. Opposition of this recommendation might request evidence that extrapolation is undesirable. We argue that this request is beside the point. Consider the laws that require a government to make all its expenditures known to the public. Transparency in this context does not mean that all spending by a government may be inappropriate or corrupt. Transparency of the expenditure allows the public and stakeholders to identify inappropriate expenses and reveal corruption. Similarly, the information about extrapolation should be made available to the stakeholders first, before any assessment of its desirability. Resistance to the transparency proposal means keeping the extrapolation information hidden.

\section{Why is it important to scrutinize extrapolation?}


In the research community, there have been discussions about whether machine learning models interpolate or extrapolate. Some researchers assume that models are predominantly interpolating between their training samples \cite{belkin2019reconciling,webb2020learning} 
and do not often extrapolate. All the datasets we have investigated prove to be extrapolating frequently enough to be taken seriously. 

On the other hand, a group of researchers recently reported that in datasets with more than 100 features, learning always amounts to extrapolation \cite{balestriero2021learning}. This notion is realistic, but two issues arise. First, it leaves out many applications where datasets have less than 100 features. Second and more importantly, this position can be used to trivialize extrapolation. Some scholars have argued that since extrapolation happens frequently, it must be trivial. Our results show the opposite. If we continue to believe that extrapolation is trivial, people affected by it may not be entitled to know about this fundamental issue.


Many applications of AI and ML are based on datasets with 10 to 50 features. Extrapolation in such applications is neither trivial nor negligible. When we studied, for example, the adult income dataset \cite{dua2017uci}, a benchmark case for studying social applications of machine learning, about half of its testing samples required some extrapolation. Some of these extrapolations may be considered negligible, but for a considerable portion of testing samples, the extent of extrapolation is far from negligible. We see the case of a woman in the US workforce originally from Thailand, with high education in a managerial position, but in the lower-income bracket. The training set of this dataset did not have any sample close to her, so significant extrapolation in the dimensions of age, native country, race, education level, and weekly work hour has to happen by any model trained on this dataset. We projected this woman's information to the convex hull of the training set and saw that in these dimensions, both collectively and individually, the projections significantly differ from hers. What we see is not just an outlier here - such levels of extrapolation are neither rare nor predominant. Consider another case in the healthcare domain. We investigated a dataset from the Veterans Affairs Healthcare System \cite{justice2006veterans} with more than one million patient records. Performing a 5-fold cross-validation, about 15\% of patient records in the testing set required extrapolation. For many of these, extrapolation was too extensive from the medical perspective to be considered proper. These trends persist in all the other datasets we studied. Extrapolation cannot be dismissed as trivial. In any respect, the affected person should have a right to know that model extrapolated when it made that decision for her.

\section{The AI Act by the European Commission}

Article 10, paragraph 2(g), of the AI Act by the European Commission requires ``identification of any possible data gaps or shortcomings." Extrapolation can be considered a ``shortcoming." Paragraph 3 then mentions that datasets should be relevant and representative, but the article does not suggest a way to quantify relevance and representativeness. Extrapolation cases may be considered a shortcoming in representation.

We suggest that the following clause be appended to article 13 of the AI Act: \textit{any decision made by an AI system should come with information on whether the model has extrapolated. If extrapolation is performed, AI systems should also report the attributes of extrapolation.}

If article 13 reveals whether a model has extrapolated or not, article 10 can be the basis to scrutinize the extrapolation information.

\section{Should we prohibit extrapolation?}

Extrapolation may lead to good and bad decisions. We do not suggest the prohibition of extrapolation. In certain applications, models may extrapolate, inevitably. Certain types of extrapolation may be justifiable by experts, while others are not. Mathematically, there are ways to ensure a model extrapolates in a desirable way in certain directions. But, for a complex model, it may not be easy to figure out how extrapolations may be acceptable beforehand. The problem is that no magic criterion would tell us when extrapolation is undesirable — determining when and where extrapolation is justified requires domain expertise. Instead of prohibition, we suggest transparency to pave the way for experts to scrutinize cases of extrapolation.




Going back to the example of transparency for expenditures of a government, it seems impossible to come up with a magic rule that would automatically identify all misappropriations and all corruptions. However, making the expenditures known to stakeholders (in this case, the public) is indeed possible. Once such information is available to the stakeholders, they can scrutinize it.



\section{Conclusions}

The community recognizes that AI and ML models may have shortcomings and unacceptable biases \cite{rudin2019stop,eshete2021making}. 
In the past two decades, data collection from various realms of life, together with growing computational power, has allowed the practice of learning from data to spread widely, leading to the emergence of a field known as data science, ML, and AI. This widespread practice can be viewed as a democratization of mathematical modeling and data analysis as researchers from one discipline often contribute to other disciplines by way of deploying AI and ML tools. 
Yet this democratization has sometimes happened at the expense of domain expertise and interpretability. Models that fall under the umbrella of AI and ML are usually complex mathematical functions that are difficult to interpret \cite{yousefzadeh2020auditing}, hence the proper name ``black-box models." Requiring explanations about the rationale behind the model's decisions has entered the public policy domain and regulations, but the knowledge about whether a model has extrapolated or not has been neglected. This shortcoming would undermine the effectiveness of the current versions of AI regulations. In the absence of AI regulations on extrapolation, individuals and civil society should consider using the existing legal system and regulations to seek transparency and scrutinize the decisions made for them by automated systems.

We shall acknowledge that AI and ML models learn from data, and the data used to train a model is finite and limited. This view empowers people who are affected by these models. They can question the model by questioning how their data compares with the data used to develop the models. Extrapolation is only one piece of explainability.

Extrapolation is not the only way a model may make bad decisions for people, and knowing about extrapolation is not the only piece of information that one needs to know about AI and ML models. Nevertheless, transparency about extrapolation would be a crucial step in regulating AI models and empowering the people affected by automated systems.

\bibliography{refs}

\end{document}